\documentclass[letterpaper, 10 pt, conference]{ieeeconf}  % Comment this line out if you need a4paper

\IEEEoverridecommandlockouts                              % This command is only needed if 
\usepackage{graphicx} % for pdf, bitmapped graphics files
\usepackage{amsmath} % assumes amsmath package installed
\usepackage{amssymb}  % assumes amsmath package installed
\usepackage{makecell}
\usepackage{hhline}
\usepackage{rotating}
\usepackage{array}
\usepackage{multirow}
\usepackage{balance}
\usepackage[utf8]{inputenc}

\title{\LARGE \bf
Large Language Models for Model-Based Robot Design}

\author{Andrew Wilhelm$^{1}$$^{*}$, Angelina Zhou$^{1}$, and Nils Napp$^{1}$% <-this % stops a space
\thanks{$^{1}$Department of Electrical and Computer Engineering, Cornell University,
Ithaca, NY, 14850, United States}%
\thanks{$^{*}$Correspondence to 
        {\tt\small ajw343@cornell.edu}}
\thanks{This material is based on work supported by the National Science Foundation grants NSF\#1846340, NSF\#2054744, and the GRFP DGE\#2139899. Any opinions, findings, and conclusions or recommendations expressed in this material are those of the author(s) and do not necessarily reflect the views of the National Science Foundation.}%
}

\begin{document}

\maketitle
\pagestyle{empty} 
% \pagestyle{plain}

%%%%%%%%%%%%%%%%%%%%%%%%%%%%%%%%%%%%%%%%%%%%%%%%%%%%%%%%%%%%%%%%%%%%%%%%%%%%%%%%
\begin{abstract}
Large Language Models (LLMs) can contribute useful engineering knowledge to robot design, but directly generated designs may rely on implicit assumptions and provide no guarantees of feasibility or optimality.
These assumptions are critical because different reasonable modeling choices can materially change which designs are predicted to be feasible or optimal.
We therefore present a framework that uses LLMs to construct explicit engineering models containing physical relationships, compatibility constraints, and objectives, allowing these modeling choices to be inspected and revised before formal optimization.
The model can then be updated with additional engineering, manufacturer, or system-specific information before formal multi-objective optimization provides feasibility and Pareto-optimality guarantees with respect to the finalized model and specified design space.
We evaluate the framework on quadcopter and line-following robot component-selection problems.
Across 30 direct LLM design trials, none could be verified as feasible under the corresponding finalized model.
Comparisons with an independently developed expert model and successive stages of model refinement further showed that changes in modeling assumptions substantially altered the predicted feasible and Pareto-optimal design sets.
Together, these results show that using LLMs to construct explicit engineering models makes the underlying design choices available for inspection and revision before those assumptions determine the optimized designs.
Explicit modeling therefore provides an interface for combining LLM-generated engineering knowledge, system-specific information, and formal design optimization.

\end{abstract}

%%%%%%%%%%%%%%%%%%%%%%%%%%%%%%%%%%%%%%%%%%%%%%%%%%%%%%%%%%%%%%%%%%%%%%%%%%%%%%%%

% \vspace{-2.5mm}
\section{INTRODUCTION}
\vspace{-1.25mm}

\noindent Model-based robot design provides a principled way to optimize complex systems by explicitly describing the physical relationships, compatibility requirements, constraints, and objectives that define a design problem.
Recent advances in model-based design have enabled efficient multi-objective optimization for complex, large-scale design spaces~\cite{wilhelmMonotoneSubsystemDecomposition2025}, allowing designers to systematically explore large design spaces and identify trade-offs among competing objectives.
However, many current model-based approaches rely on expertly crafted models, requiring a designer to translate high-level design requirements and engineering knowledge into a formal representation before optimization can begin.

\begin{figure}[t]
      \centering
      
      \includegraphics[scale=0.525]{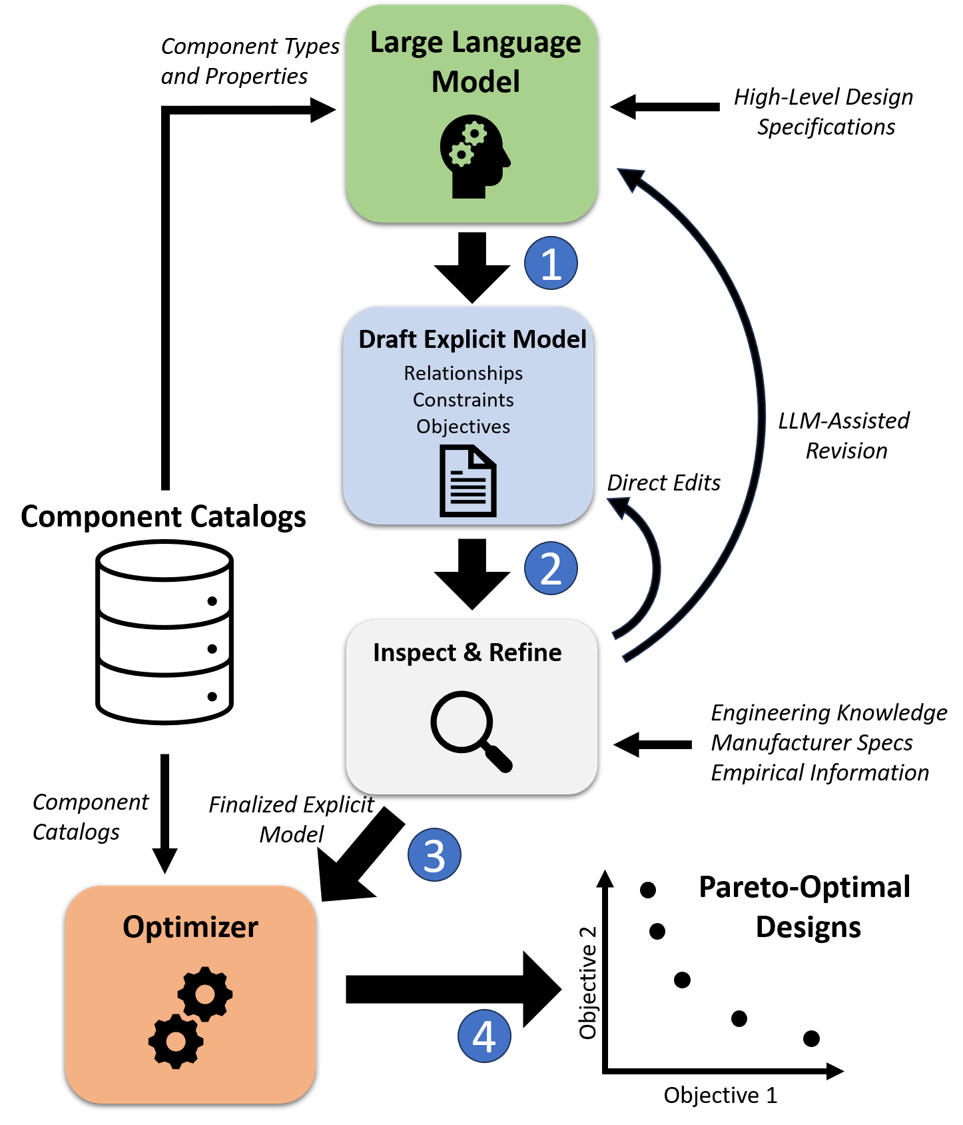}
      \vspace{-5mm}
      
      \caption[Caption for LOF]{\small Overview of the proposed framework for LLM-assisted model-based robot design. (1) High-level design specifications and grounded component information are provided to the LLM, which drafts an explicit engineering model. (2) The model is inspected and revised to incorporate additional engineering or system-specific information as needed. (3) The finalized model and component catalogs are passed to a formal optimizer, which (4) determines Pareto-optimal designs with respect to the stated model and design space.
}

      \label{LLMRobotDesignPipelineFig}
    \vspace{-5mm}
  \end{figure}

Large Language Models (LLMs) are well suited to assist with this modeling process.
LLMs excel at synthesizing information, and many robot design models are built from physics-based relationships and compatibility constraints that can be described in natural language and expressed mathematically.
Given high-level design requirements and information about the available design variables, an LLM can help identify relevant relationships, constraints, intermediate quantities, and objectives, and can revise the model through natural-language interaction.

While LLMs are promising tools for constructing robot design models, relying on their reasoning directly for robot design introduces several critical challenges, particularly for tasks involving geometric reasoning, multi-step numerical computation, and optimization under conflicting constraints~\cite{mysiorEvaluatingLargeLanguage2026}.
LLM outputs may be ungrounded, leading to hallucinations or designs that overlook important interactions between components and system-level requirements~\cite{makaturaLargeLanguageModels2024}.
Moreover, even when an LLM identifies the appropriate physical relationship, it may not know the correct system-specific parameters, such as friction or drivetrain losses, which may require experimental characterization rather than being inferred from general engineering knowledge or component specifications alone.
Making these assumptions explicit allows a designer to inspect, correct, or replace them with system-specific information.
Finally, LLM-generated designs inherently lack guarantees of feasibility or optimality: a plausible design does not establish that all modeled constraints have been simultaneously satisfied or that a better design does not exist.
Thus, stronger LLM reasoning and formal model-based optimization provide complementary capabilities rather than interchangeable ones.

Our primary contribution is a framework for integrating LLM reasoning with explicit model-based robot design, in which the LLM helps construct the design model rather than directly selecting a final design (Fig.~\ref{LLMRobotDesignPipelineFig}).
The LLM receives high-level design requirements together with grounded information about the available design variables and their properties, and generates a structured model containing relationships, constraints, and objectives.
A designer then inspects and iteratively refines the model, correcting assumptions and incorporating additional engineering or empirical knowledge where needed.
The resulting model interfaces with a formal optimizer, which solves the multi-objective design problem and returns feasible, Pareto-optimal designs with respect to the stated model.
In this way, the explicit model provides a common interface between LLM-generated engineering knowledge, designer expertise, empirical characterization, and formal optimization.

While the framework applies more broadly to model-based robot design, in this work we evaluate it on system-level component-selection problems for a mini quadcopter and a differential-drive, line-following robot, with design spaces containing over $7.73\cdot10^{10}$ and $2.18\cdot10^{13}$ possible component combinations, respectively.
We compare our framework against an end-to-end LLM baseline, in which the LLM directly selects components from the same catalogs using the same design requirements.
For the line-following robot, we compare the final LLM-assisted model against an independently developed expert-authored model.
We additionally examine how the design space changes during the actual iterative model-refinement process and investigate model disagreements using physical or independent engineering evidence.
Together, these experiments show that LLMs can contribute to robot design by constructing explicit engineering models that expose the assumptions underlying the design problem before those modeling choices shape the feasible and Pareto-optimal designs produced by formal optimization.

% \vspace{-2mm}
\section{RELATED WORK}
\label{sec:related-work}

\subsection{Model-Based Robot Design and Optimization} 

Automated robot design includes methods that generate and evaluate candidate robot designs, as well as model-based methods that encode the design problem in an explicit mathematical representation for optimization.
Generative approaches have used evolutionary algorithms~\cite{bhatiaEvolutionGymLargeScale2021} and grammar-based representations~\cite{zhaoRoboGrammarGraphGrammar2020} to co-design robot morphologies and controllers.
These methods are particularly useful for exploring broad or poorly characterized design spaces without requiring the relevant engineering relationships to be represented in a unified analytical model.

In model-based design, the relationships and constraints among system elements and design variables are explicitly modeled.
While specifying an exact robot model restricts the design space, it enables the use of classical optimization techniques for optimizing the results, including Monotone Co-Design Problems~\cite{censiMathematicalTheoryCoDesign2015} and constrained optimization techniques such as integer~\cite{magnussenMulticopterDesignOptimization2015} or quadratic programming~\cite{haComputationalCooptimizationDesign2018}.
For component-level robot design in particular, integer~\cite{carloneRobotCodesignMonotone2019} and constraint~\cite{wilhelmConstraintProgrammingComponentLevel2023} programming have been used to select compatible components while satisfying physical and performance requirements.

These model-based approaches provide an explicit representation through which physical relationships and compatibility requirements can be enforced and competing objectives can be optimized, but they generally assume that the engineering model has already been specified.
Our work addresses this preceding model-construction stage: rather than replacing formal optimization with an LLM, we use an LLM to assist in constructing the explicit relationships, constraints, and objectives that define the robot design problem before it is solved using an optimization method.

\subsection{LLMs for Robot and Engineering Design}
The rapid development of LLMs has motivated their use throughout robotics and engineering design, including for developing robot applications through natural-language interaction~\cite{karliAlchemistLLMAidedEndUser2024} and high-level task planning~\cite{huangInnerMonologueEmbodied2022}.
In robot design specifically, recent methods have incorporated LLMs into generative design pipelines.
RoboMorph~\cite{qiu2026robomorph} uses an LLM as a generative operator within an evolutionary loop to develop robot morphologies that are subsequently evaluated using learned controllers.
Similarly, LASeR~\cite{songLaserDiversifiedGeneralizable2025} uses LLM-generated proposals within an evolutionary robot-design process, where evaluated designs from previous generations inform subsequent generation.
These methods demonstrate how LLMs can contribute engineering priors and generative capabilities while downstream evaluation and optimization determine which candidate designs perform well.

LLMs have also been investigated for more interactive and direct forms of engineering design.
Stella et al.~\cite{stellaHowCanLLMs2023} propose a general framework for robot design, where a user starts with high-level questions about possible robot designs and the design space is iteratively narrowed through interaction with the LLM.
Makatura et al.~\cite{makaturaLargeLanguageModels2024} performed an extensive analysis of LLMs across several stages of the design and manufacturing pipeline.
Of particular relevance to system-level robot design, their ``Part Sourcing'' experiments ask an LLM to identify and select components for physical systems including a quadcopter.
They identified several challenges with their approach, with the LLM selecting redundant components, hallucinating components that did not exist, providing incorrect specifications of components, and selecting incompatible components.
We use a related direct component-selection approach as the end-to-end LLM baseline in our experiments, while providing the LLM with the same fixed component catalogs available to our model-based pipeline.

These approaches primarily place the LLM within the design process to generate or select candidate designs that are subsequently evaluated.
Our approach instead uses the LLM before design optimization to help construct an explicit engineering model of the design problem.
The physical relationships, compatibility requirements, constraints, and objectives represented in this model can then be evaluated and revised independently of the LLM before formal optimization determines the final designs.

\subsection{LLM-Assisted Model Construction and Optimization}
A related line of work has investigated using LLMs to construct mathematical optimization models from natural-language problem descriptions~\cite{ahmedLM4OPTUnveilingPotential2024}.
OptiMUS~\cite{ahmaditeshniziOptiMUS2024}, for example, uses an LLM-based system to formulate mixed-integer linear programs from natural-language descriptions, generate solver code, evaluate generated solutions, and revise formulations when necessary.
More recently, Astorga et al.~\cite{astorgaAutoformulationMathematical2025} study the automated creation of solver-ready optimization models, treating formulation as a structured search problem and using LLM-based evaluation together with symbolic pruning to explore candidate formulations.

These works demonstrate that LLMs can help translate problem descriptions into formal optimization representations.
The engineering modeling problem considered here is related, but importantly different: high-level robot design requirements generally underdetermine the mathematical model itself.
They do not uniquely specify which physical relationships should be modeled, which compatibility requirements are necessary, what approximations are appropriate, or which system-specific parameters should be used.
Multiple physically reasonable formulations may therefore satisfy the same high-level description and lead to different feasible or optimal designs.
Moreover, some information required to construct an adequate model, such as component-specific compatibility details, drivetrain losses, friction, or other system-specific dynamics and parameters, may not be present in the high-level problem description or general engineering knowledge and may instead require manufacturer data or empirical characterization.
This limitation is therefore not solely a matter of improving LLM reasoning: some information necessary to finalize the engineering model is inherently problem- and system-specific.

Our work builds on these prior directions by using the LLM to help construct the explicit engineering model rather than to directly propose or select candidate robot designs.
The resulting model can be examined and updated with problem- and system-specific information before being passed to a formal optimizer, which performs the design optimization and provides feasibility and Pareto-optimality guarantees with respect to the finalized model and specified design space.

% \vspace{-2mm}
\section{METHODS}
\label{sec:methods}

In this section, we present our framework for using LLMs to assist with explicit model-based robot design. 
The process begins with high-level design requirements and grounded information about the available design variables, which are provided to an LLM to generate an explicit model. 
A designer then inspects and refines this model before it is passed to a formal optimizer. 
We first describe this general modeling pipeline and then present the component-selection formulation used to instantiate and evaluate it in this work.

\subsection{LLM-Assisted Robot Design using Explicit Models}
Our framework uses the LLM to construct an explicit robot design model rather than directly select a final design.
Given high-level design requirements and grounded information about the available design variables, the LLM generates a model containing relevant variables, intermediate physical relationships, constraints, and objectives.
By expressing these design considerations in an explicit model, assumptions made during model generation can be examined and revised before optimization.

The initial model is then iteratively reviewed and refined by the designer.
The designer can directly modify the model or interact with the LLM to correct equations or assumptions, add missing physical or compatibility relationships, and replace simplified models where greater fidelity is required.
The designer can also incorporate system-specific or empirical information that may not be reliably inferred from general engineering knowledge or component specifications alone.

Once the model has been reviewed and finalized, it is passed to a formal optimization method that searches the specified design space subject to the model constraints.
For multi-objective problems, the optimizer determines the Pareto-optimal designs under the stated objectives.
Feasibility and Pareto optimality are guaranteed with respect to the finalized explicit model and specified design space, not necessarily with respect to the underlying physical system.

In the following sections, we describe how we implement this framework for the component-selection problems considered in this work.

\subsection{Model Generation and Iterative Refinement} 

For the component-selection problems considered in this work, the available design variables and their properties are specified by structured component catalogs.
In the initial prompt, we provide the LLM with the high-level robot design specifications and the component types available for the design problem.
We use a retrieval-augmented generation (RAG) system~\cite{lewisRetrievalAugmentedGenerationKnowledgeIntensive2020} to query the component database and retrieve, for each component type, a list of its properties, their respective units, and the frequency with which each property is populated.
This grounds the generated model in properties that are actually available in the component catalogs.
We also instruct the LLM to prioritize well-populated properties, since the optimizer can only evaluate components for which all referenced properties are available.
Finally, we provide the LLM with an example model using one-shot prompting to demonstrate the desired structure of the output.

Using this information, the LLM generates a preliminary model in a structured format that can be processed by the downstream optimization pipeline.
The model uses the component types, property names, and units supplied from the catalogs and defines the intermediate quantities, physical and compatibility relationships, equality and inequality constraints, and objectives for the design problem.

The designer then reviews the generated model and either modifies expressions directly or asks the LLM to revise them.
During this process, the designer can request corrections to equations, additional physical or compatibility constraints, or more detailed representations of particular subsystems.
For example, the designer may add a missing shaft-diameter compatibility constraint, ask the LLM to reformulate a motor model, or request that effects such as friction or drivetrain losses be included.
The designer can also ask the LLM to explain parts of the model or suggest additional design considerations that may have been omitted.
Measured parameters or other system-specific information can be added directly to the model when they cannot be reliably determined from general engineering knowledge or component specifications.

The resulting finalized model is then represented and solved using the component-selection formulation described below.

\subsection{Component-Selection Formulation and Formal Optimization} 
\label{notation_subsection}

In this work, we evaluate the framework on system-level component-selection problems. 
Given a robot model and catalogs of available components, the goal is to select one component for each required component type such that the resulting design satisfies the modeled physical, performance, and compatibility constraints while optimizing one or more design objectives. 
Because the number of possible component combinations grows rapidly with catalog size, we formulate and solve this problem using constraint programming.

% The framework described above is not tied to a particular optimization method.
For the component-selection problems in this work, we use the constraint-programming formulation from \cite{wilhelmMonotoneSubsystemDecomposition2025}; we summarize the notation here and refer to that work for the complete formulation.

The component selection problem in robot design can be naturally expressed as a constrained optimization problem.
The goal is to determine variable assignments from their respective domains that optimize the design objectives while satisfying all constraints.
Formally, we define a finite index set of variables $V=\{1,2,\ldots,N_V\}$, each associated with a finite domain $D_v\in\mathcal{D}=\{D_1,\ldots,D_{N_V}\}$, and a finite index set of constraints $C={1,2,\ldots,N_C}$.
Each element $d_v\in D_v$ is a tuple representing an available catalog entry, where the tuple entries contain the properties associated with that component type.
The $p$-th property of a variable $v$ is denoted as $D_{v,p}$.
For component selection, each variable corresponds to a specific component type, its domain consists of available catalog entries, and the associated properties $D_{v,p}$ define the physical characteristics of those components and are used to establish relationships with other components.
For example, in the line-following robot problem, a variable $v$ may correspond to the selection of a wheel and is characterized by properties such as mass, diameter, and cost.

Constraints are mathematical expressions describing relationships between component properties.
These not only enforce valid combinations but also encode performance specifications.
Inequality constraints take the general form $L_c(D_{v,p},D_{v',p'},\ldots)\leq R_c(D_{v,p},D_{v',p'},\ldots),$ where $L_c$ and $R_c$ are real-valued functions over the variable properties that define constraint $c\in C$; equality constraints are defined analogously.
Similarly, objective functions are defined on variable properties, $f_i(\mathbf{v})\equiv f_i(D_{v,p},D_{v',p'},\ldots)$, and may be minimized or maximized.

For readability, we associate the index-based variables and properties with descriptive symbols.
For example, if $M$ denotes the motor variable and $W$ the wheel variable, we denote the selected motor's rotational-speed property as $M_{\textrm{rps}}$ and the selected wheel's diameter as $W_{\textrm{diameter}}$.
A minimum-speed constraint can then be written as $target\_speed \leq M_{\textrm{rps}} W_{\textrm{diameter}} \pi,$ where $target\_speed$ is a high-level design constant specifying the robot's desired nominal speed, and the right-hand side gives the linear speed corresponding to the selected motor and wheel.
We use the LLM to generate such constraint expressions for a given robot design.
Representing these relationships individually makes the physical, performance, and compatibility assumptions in the model explicit so they can be inspected and revised before optimization.

Once the problem is formulated in this way, we use IBM ILOG CP Optimizer, part of IBM ILOG CPLEX Optimization Studio, to solve the resulting constraint-programming problem.
For the multi-objective problems considered in this work, the optimization procedure computes a Pareto front of optimal component selections, representing trade-offs among the competing objectives.

\subsection{Experimental Setup}

We evaluate the proposed framework on two case studies: designing a quadcopter and a line-following robot.
The quadcopter problem contains eight component types with 33--96 candidate components per type, while the line-following robot contains seven component types with 17--491 candidates per type.
We use GPT-5.5 through the OpenAI API for all LLM-assisted model generation and refinement, as well as for the end-to-end LLM baseline described below.
The finalized quadcopter model contains 8 component-selection variables, 34 constraints, and 3 objectives, while the finalized line-following robot model contains 7 component-selection variables, 61 constraints, and 3 objectives.
A representative excerpt of the finalized quadcopter model is provided in the Appendix.
For both problems, the finalized LLM-assisted model is solved over the corresponding component catalogs to obtain a Pareto front of optimal designs, and we compare the results against an end-to-end LLM baseline.
For the line-following robot, we additionally compare the LLM-assisted model with an independently developed expert-authored model and analyze how iterative model refinement changes the resulting designs.

To contextualize performance, we use a baseline approach in which the same LLM directly selects components to meet the high-level design specifications and optimize for the performance objectives, based on the end-to-end LLM part-sourcing approach used in \cite{makaturaLargeLanguageModels2024}.
For both robot problems, the LLM is given the same high-level requirements and component catalogs as our framework, but is not given the explicit constraint model or access to the optimization solver.
Because the baseline produces a single design for a specified objective rather than explicitly solving the multi-objective problem, we run it independently five times for each objective.
Each run is conducted in a separate conversation with no shared context between trials, using the same input information except for the specified optimization objective.
Each generated design is then evaluated using the finalized explicit model to determine feasibility, violated constraints, objective values, and, for feasible designs, whether it is dominated by designs on the corresponding Pareto front.

For the line-following robot, we additionally compare the final LLM-assisted model against an independently developed expert-authored model. 
The expert model was developed without access to the LLM-assisted model and is treated as an independent engineering reference rather than ground truth.
We solve both models using the same component catalogs and design requirements, then evaluate the Pareto-optimal designs from each model under the other model.
This cross-evaluation identifies designs that are feasible under both formulations as well as disagreements arising from differences in the modeled assumptions and constraints.

Finally, we use model versions from the actual model-refinement history to test whether iterative refinement materially changes the designs considered feasible or optimal, and whether designs produced by an earlier version of the model remain feasible or optimal under the finalized formulation.
We examine representative intermediate versions of the LLM-assisted model, solve each independently, and evaluate their Pareto-optimal designs under the finalized model.
This allows us to identify which later model revisions invalidate earlier solutions and how refinement changes the regions of the design space considered feasible or optimal. 
Where useful for interpreting differences between models or stages of refinement, we examine the underlying constraints and modeling assumptions, including system-specific information incorporated during model refinement.

% \vspace{-2mm}
\section{Results}
% \vspace{-1mm}
\label{results}
% \vspace{-2.5mm}

\subsection{End-to-End LLM Design Reliability}

\begin{table}[t]
    \vspace{3mm}
    \centering
    \caption{\small End-to-end LLM baseline results.}
    \label{tab:end_to_end_baseline}
    \small
    \setlength{\tabcolsep}{1.75pt}
    \renewcommand{\arraystretch}{1.05}
    \begin{tabular}{clccc}
        \hline
        Robot & Objective & Feasible & Violated & Indeterminate \\
        \hline
        \multirow{3}{*}{Quadcopter}
            & Min. cost     & 0/5 & 3/5 & 2/5 \\
            & Max. velocity & 0/5 & 3/5 & 2/5 \\
            & Max. payload  & 0/5 & 3/5 & 2/5 \\
        \hline
        \multirow{3}{*}{\makecell[c]{Line-Following\\Robot}}
            & Min. cost     & 0/5 & 5/5 & 0/5 \\
            & Max. velocity & 0/5 & 5/5 & 0/5 \\
            & Max. payload  & 0/5 & 4/5 & 1/5 \\
        \hline
    \end{tabular}

\vspace{-3mm}
\end{table}

\begin{figure*}[t]
    \centering

    \begin{minipage}[t]{0.48\textwidth}
        \centering
        \includegraphics[width=\linewidth]{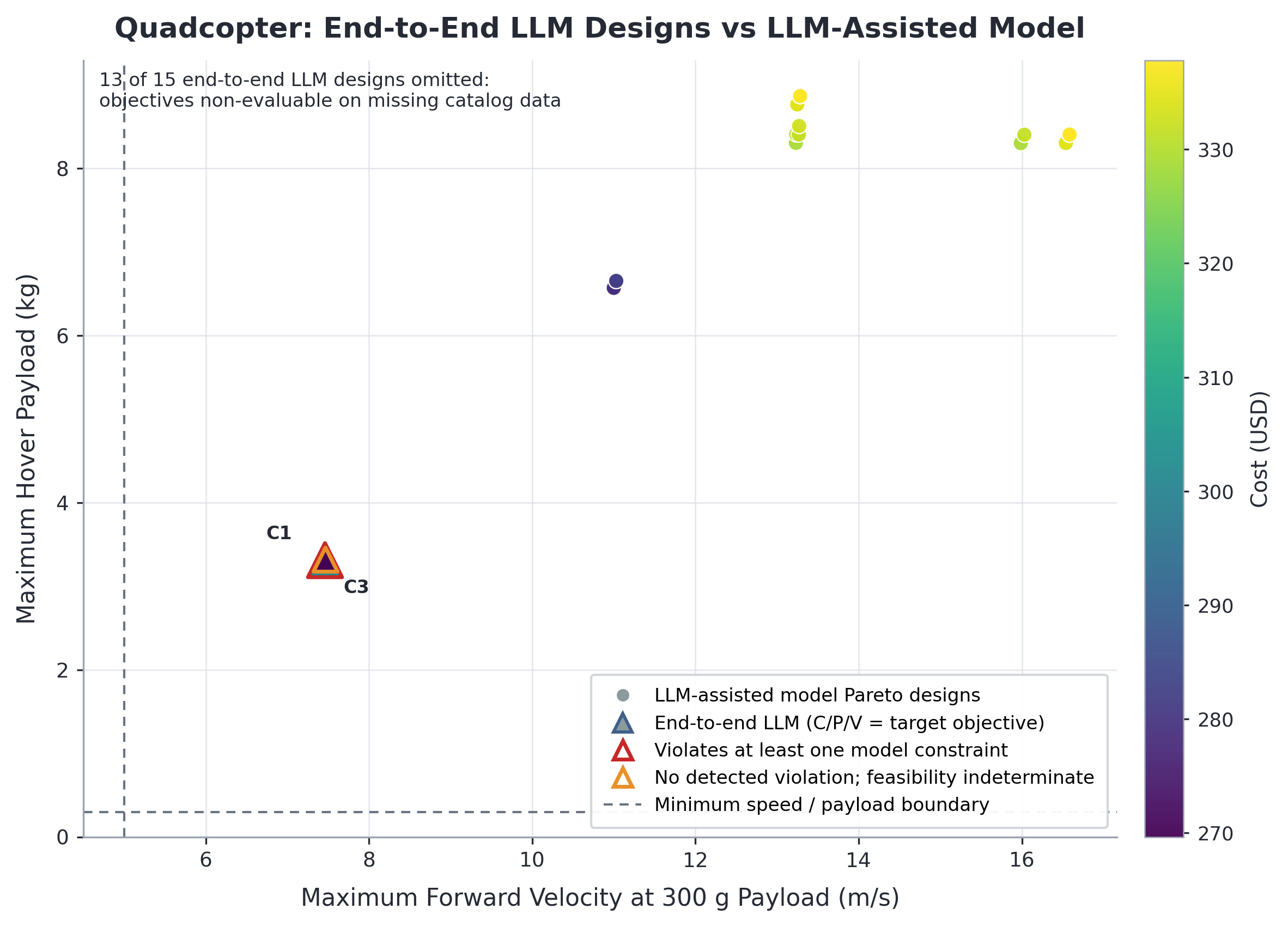}
        
        \vspace{-1mm}
        \small (a) Quadcopter
    \end{minipage}
    \hfill
    \begin{minipage}[t]{0.48\textwidth}
        \centering
        \includegraphics[width=\linewidth]{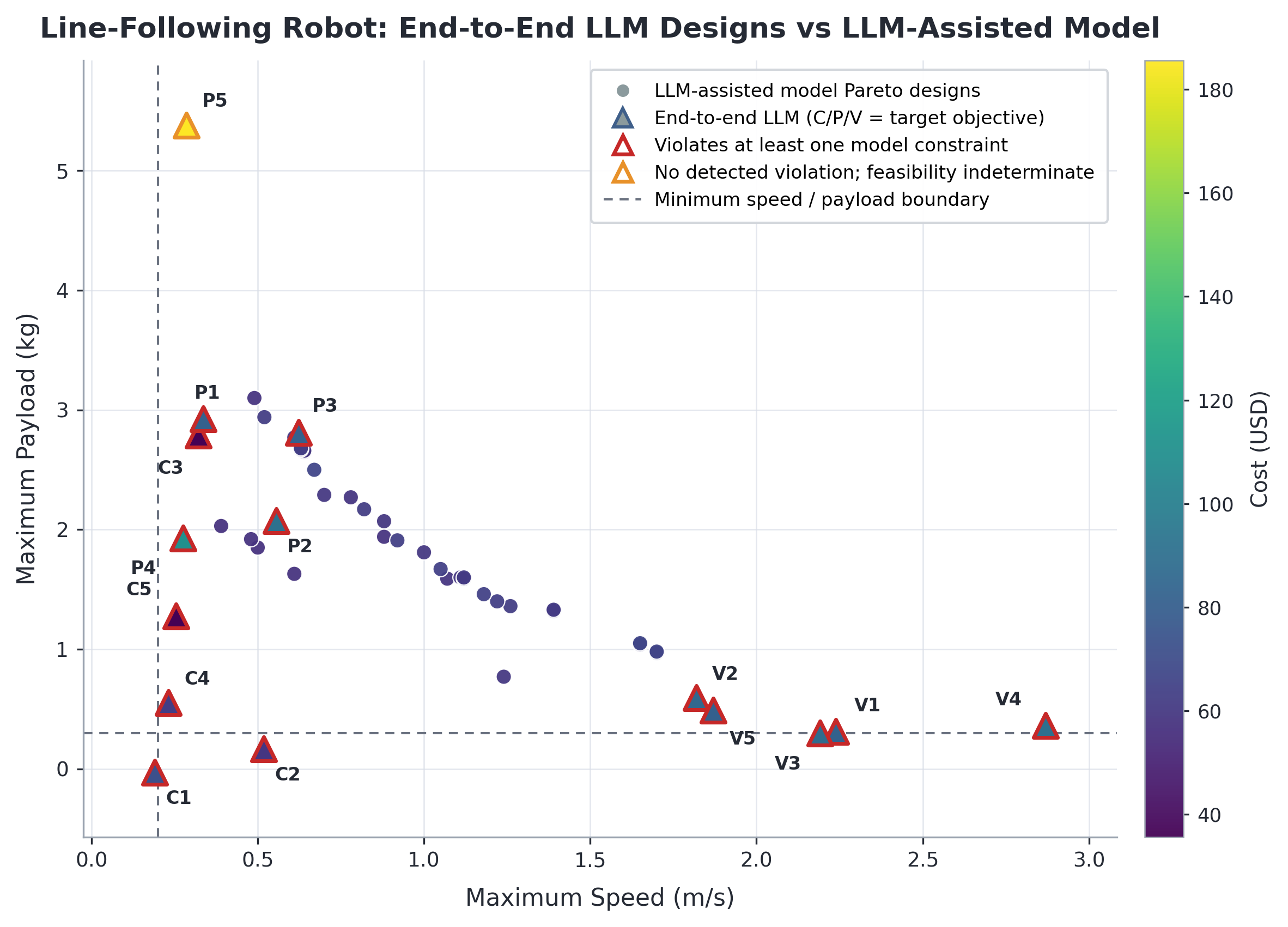}
        
        \vspace{-1mm}
        \small (b) Line-following robot
    \end{minipage}

    \caption{\small
    End-to-end LLM designs compared with the Pareto-optimal designs obtained using the finalized explicit models for (a) the quadcopter and (b) the line-following robot. The LLM was queried independently five times for each optimization objective: minimum cost, maximum velocity, and maximum payload. Red outlines indicate designs that violate at least one evaluable constraint, while orange outlines indicate designs with no detected violation whose full feasibility cannot be determined because required component properties are missing. Thirteen of the fifteen quadcopter designs are omitted because missing catalog properties prevent evaluation of the plotted objectives.
    }
    \label{fig:end_to_end_baseline}
    % \vspace{-2mm}
\end{figure*}

\noindent We first evaluate whether an LLM can reliably perform component selection directly from the same high-level design requirements and component catalogs used by our explicit model-assisted pipeline, but doing so without access to the explicit model or optimization solver.
To contextualize performance, we compare the resulting designs against the Pareto-optimal solutions obtained using the finalized explicit models for the quadcopter and line-following robot design problems.
Across five independent runs for each of the three objectives, the end-to-end LLM generated a total of fifteen designs for each robot problem.

Table~\ref{tab:end_to_end_baseline} summarizes the feasibility of these solutions.
Across all 30 baseline runs, none could be verified as feasible under the corresponding finalized model.
Twenty-three designs violated at least one evaluable constraint, while the remaining seven could not be fully evaluated because required component properties were missing.
Missing properties were particularly common for the quadcopter designs. 
Most undefined objective values arose because the finalized model uses motor continuous power to calculate sustainable payload and power-limited forward velocity, while many LLM-selected motors lacked this property. 
The LLM could instead reason from other available specifications or approximations, such as peak motor power, but these are not interchangeable with continuous power for sustained operation.
Thus, despite receiving the same component catalogs and design requirements as the model-based pipeline, direct LLM component selection did not reliably produce designs for which all modeled physical, performance, and compatibility requirements could be verified.

Figure~\ref{fig:end_to_end_baseline} shows the end-to-end LLM designs for which the plotted objective values could be evaluated, together with the Pareto fronts obtained using the finalized explicit models. 
These results show that the LLM responds to the specified optimization objective and produces substantial run-to-run variation, but favorable objective values do not imply that the coupled design constraints are simultaneously satisfied.

Across the repeated baseline runs, design violations were not confined to a single subsystem. 
Line-following robot designs violated motor operating-point, electrical, endurance, and payload requirements, while quadcopter failures primarily involved motor and ESC power requirements. 
These results indicate that direct LLM component selection can identify promising components but does not reliably satisfy the full set of coupled design requirements simultaneously.

\subsection{LLM-Assisted vs. Expert-Authored Models} 

\begin{figure*}[t]
    \centering

    \begin{minipage}[t]{0.48\textwidth}
        \centering
        \includegraphics[width=\linewidth]{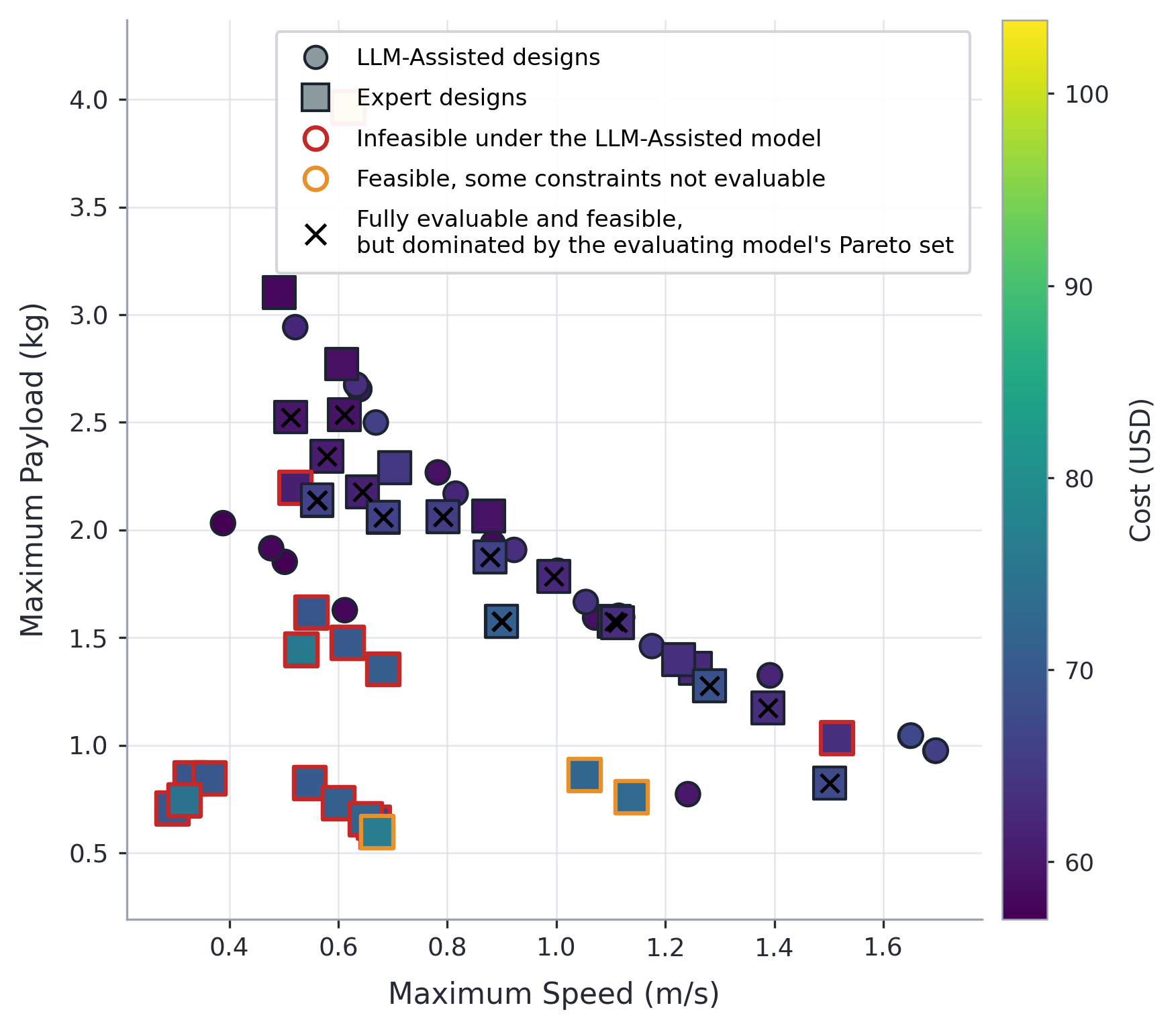}

        \vspace{-1mm}
        \small (a) Evaluated under LLM-assisted model
    \end{minipage}
    \hfill
    \begin{minipage}[t]{0.48\textwidth}
        \centering
        \includegraphics[width=\linewidth]{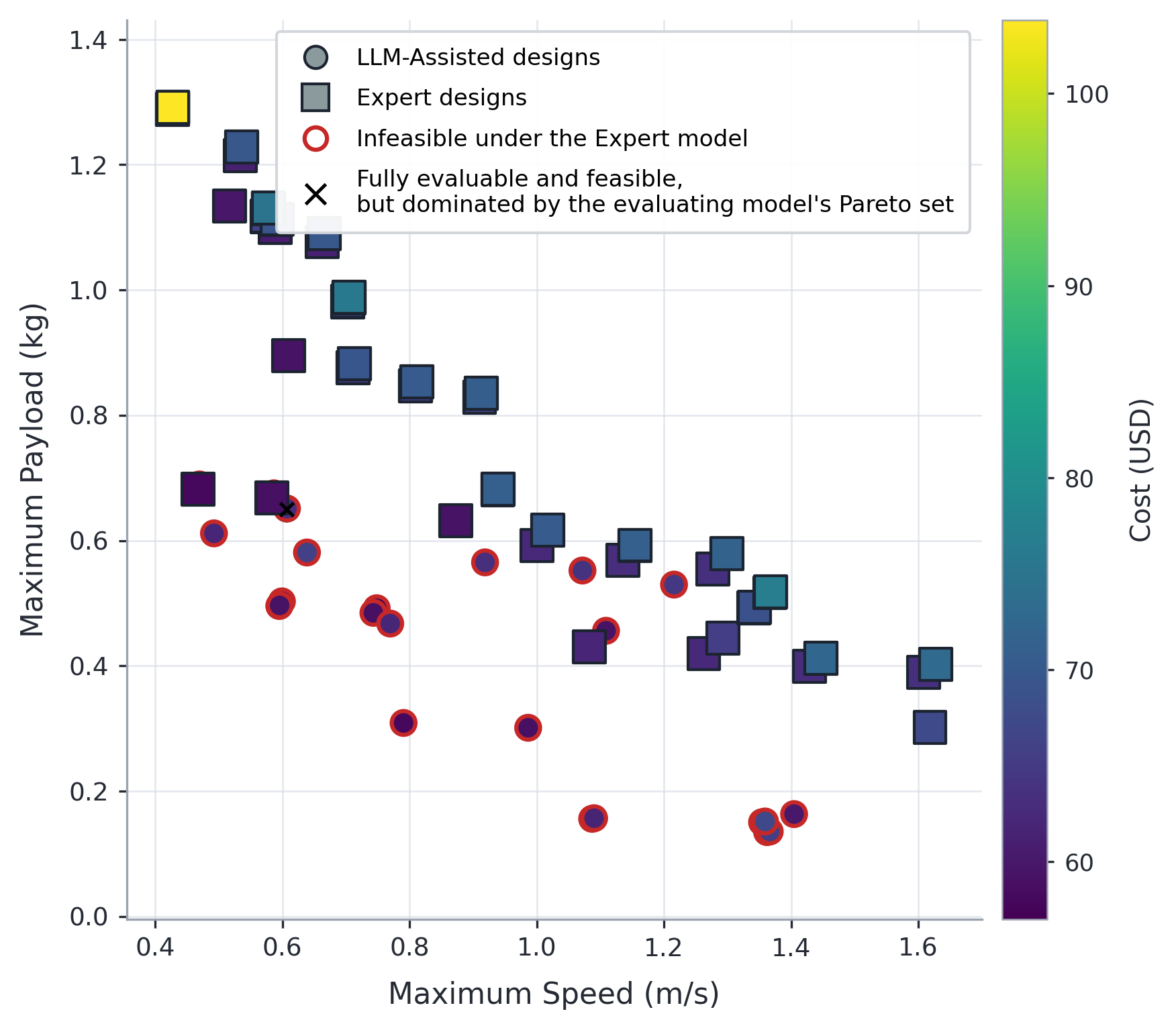}

        \vspace{-1mm}
        \small (b) Evaluated under expert-authored model
    \end{minipage}

    \caption{\small
    Cross-evaluation of Pareto-optimal line-following robot designs from the finalized LLM-assisted and expert-authored models. (a) Both design sets evaluated under the LLM-assisted model; (b) both evaluated under the expert-authored model. Red outlines indicate infeasible designs, orange outlines indicate indeterminate feasibility due to missing properties, and crosses indicate feasible designs dominated by the evaluating model's Pareto set.
    }
    \label{fig:model_cross_evaluation}
    \vspace{-3mm}
\end{figure*}

We next compare the finalized LLM-assisted line-following robot model with an independently developed expert-authored model of the same design problem.
Both models use the same component catalogs and high-level design requirements, but were developed independently and may encode different relationships, constraints, and objective definitions.
The expert model is not a ground truth but instead provides a separate engineering formulation against which we can compare the LLM-assisted model.

Figure~\ref{fig:model_cross_evaluation} compares the two Pareto sets after reevaluating them under the other model. 
Under the LLM-assisted model (Fig.~\ref{fig:model_cross_evaluation}(a)), 29 of the 69 expert-model Pareto designs (42.0\%) satisfy all constraints, while 6 additional designs satisfy all evaluable constraints but cannot be fully classified because several voltage-regulator constraints depend on an unavailable component property. 
Of the 29 fully evaluable and feasible expert designs, 22 are dominated by the LLM-assisted model's Pareto set and 7 remain non-dominated. 
In the reverse comparison (Fig.~\ref{fig:model_cross_evaluation}(b)), 7 of the 34 LLM-assisted Pareto designs (20.6\%) satisfy all expert-model constraints; of these, only one is dominated by the expert-model Pareto set. 
Thus, although the two models differ substantially in which designs they consider feasible, several designs remain competitive when evaluated under the other formulation.

\begin{figure*}[t]
    \centering

    \begin{minipage}[t]{0.48\textwidth}
        \centering
        \includegraphics[width=\linewidth]{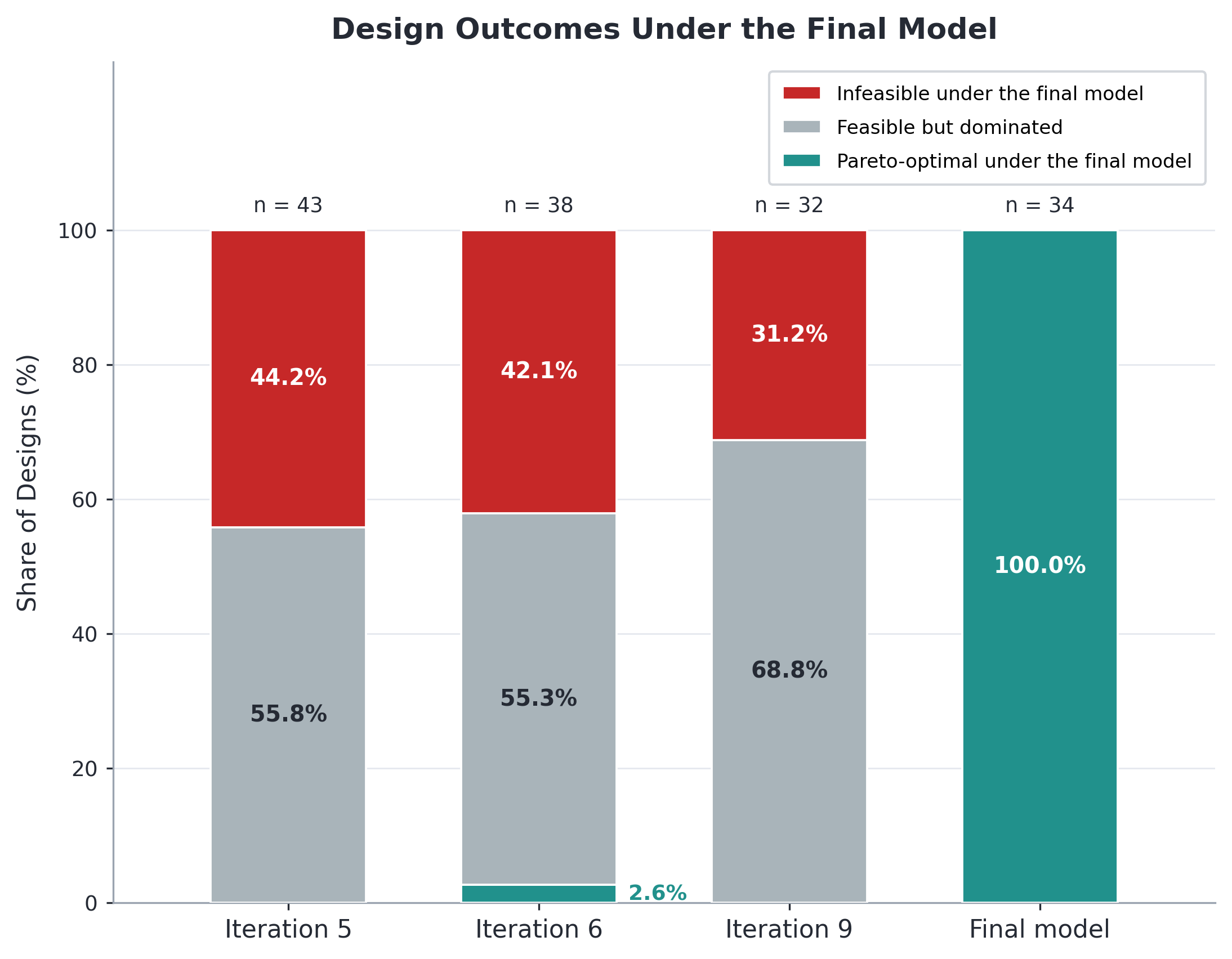}

        \vspace{-1mm}
        \small (a) Designs evaluated under the finalized model
    \end{minipage}
    \hfill
    \begin{minipage}[t]{0.48\textwidth}
        \centering
        \includegraphics[width=\linewidth]{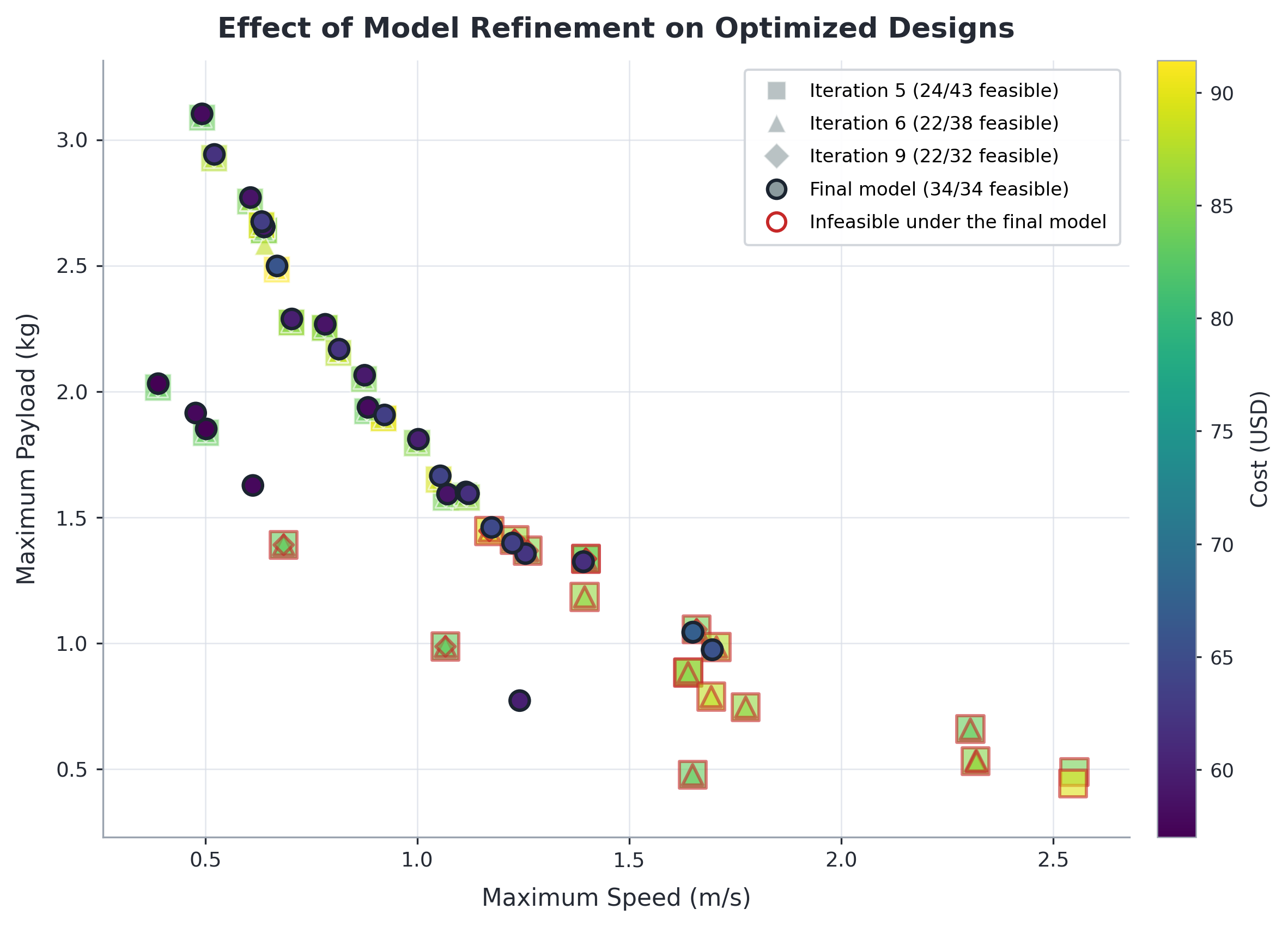}

        \vspace{-1mm}
        \small (b) Designs in the finalized model objective space
    \end{minipage}

    \caption{\small
    Effect of iterative model refinement on line-following robot design outcomes. We compare the earliest model version in the refinement history containing the full three-objective formulation (Iteration 5), two later intermediate versions, and the finalized model. (a) Pareto-optimal designs produced by each version are reevaluated under the finalized model and classified as infeasible, feasible but dominated, or Pareto-optimal. (b) The same designs are shown after recomputing their objective values using the finalized model, together with the finalized Pareto front.
    }
    \label{fig:model_refinement}
\end{figure*}

Cross-model infeasibility primarily resulted from differences in the assumptions encoded by the two models. 
Battery endurance is the largest source of disagreement: the LLM-assisted model includes efficiency losses and assumes only a fraction of listed battery energy is usable, whereas the expert model uses different power calculations and the full listed battery energy. 
The models also represent the power architecture differently, leading to different motor-supply compatibility requirements. 
These disagreements do not establish either model as ground truth; rather, they expose specific engineering assumptions that can be inspected and revised using system-specific evidence.

\subsection{Evaluating Iterative Model Refinement}

We next examine how the designs produced by early versions of the LLM-assisted model change as the model is refined, and whether those revisions substantially affect which designs are considered feasible or Pareto-optimal.
For the line-following robot, we compare four versions of the model from the actual refinement process: Iteration 5, the first version containing all three objectives; Iteration 6, which adds drag at the rear contact point; Iteration 9, which uses more conservative values for empirical constants; and the finalized model, which incorporates additional system-specific updates based on prior physical observations.
Each model is solved independently to obtain its Pareto-optimal designs, and the solutions generated at each stage are then reevaluated using the constraints and objective definitions of the finalized model.

Figure~\ref{fig:model_refinement}(a) summarizes how the Pareto-optimal designs generated by each model are classified under the finalized model.
For the earliest evaluated model, Iteration 5, 24 of 43 designs (55.8\%) remain feasible under the finalized model, while 19 (44.2\%) violate one or more finalized-model constraints. 
However, all 24 feasible designs are dominated by the finalized Pareto front. 
Feasibility increases across later model versions, reaching 22 of 32 designs (68.8\%) by Iteration 9, although nearly all feasible designs from the intermediate models remain dominated under the finalized formulation; only one Iteration 6 design remains Pareto-optimal.

Figure~\ref{fig:model_refinement}(b) shows these designs after reevaluating their objective values using the finalized model.
The earlier model versions generate designs across substantially different regions of the objective space, including designs that later become infeasible and others that remain feasible but are outperformed by solutions from the finalized model. 
Thus, refinement changes not only which designs satisfy the modeled constraints, but also which feasible designs are considered optimal.
These changes result from revisions to the physical relationships, compatibility constraints, and system-specific parameters represented in the model. 
More broadly, the results show why making the LLM-generated model explicit is important: assumptions introduced during model generation can propagate directly to the designs selected by the optimizer unless they are checked against additional engineering or system-specific information.

% \vspace{-2mm}
\section{CONCLUSION}
\label{conclusion}
\vspace{-1mm}

\noindent 
In this work, we present a framework for using LLMs to assist with explicit model-based robot design.
Rather than directly selecting a final design, the LLM helps construct a structured model whose relationships, constraints, and objectives can be inspected and revised with additional engineering or system-specific information before formal optimization.
Across our two case studies, none of the 30 designs produced by the direct LLM baseline could be verified as feasible under the corresponding finalized model.
Comparison with an independently developed expert model further showed that different engineering assumptions can lead to substantially different feasible and Pareto-optimal design sets, while iterative refinement substantially changed which designs were considered feasible or optimal.
Together, these results show that making the design model explicit provides more than an interface to an optimizer: it exposes the assumptions that determine design outcomes and allows them to be examined, compared, and revised before formal optimization.

Although we evaluate the framework here on system-level component selection, the same principle applies more broadly to robot design problems that can be represented with an explicit model and solved using formal optimization.
Future work can extend the framework beyond component selection, improve automated checking of LLM-generated models, and incorporate additional system-specific information.

% \section*{ACKNOWLEDGMENT}
% The authors utilized generative AI tools (Gemini, Claude, and ChatGPT) to assist with editing the manuscript (Secs. I–VI), as well as with code generation, data analysis, and figure plotting for the experimental results in Sec. IV. 
% All AI-assisted content, analyses, code, and figures were reviewed and verified by the authors.
\section*{ACKNOWLEDGMENT}
Generative AI tools (Gemini, Claude, and ChatGPT) were used for manuscript editing and to assist with code generation, data analysis, and figure plotting. All AI-assisted content and analyses were reviewed and verified by the authors.

% \balance

%%%%%%%%%%%%%%%%%%%%%%%%%%%%%%%%%%%%%%%%%%%%%%%%%%%%%%%%%%%%%%%%%%%%%%%%%%%%%%%%

%%%%%%%%%%%%%%%%%%%%%%%%%%%%%%%%%%%%%%%%%%%%%%%%%%%%%%%%%%%%%%%%%%%%%%%%%%%%%%%%

%%%%%%%%%%%%%%%%%%%%%%%%%%%%%%%%%%%%%%%%%%%%%%%%%%%%%%%%%%%%%%%%%%%%%%%%%%%%%%%%
% \vspace{-2mm}
% \section*{ACKNOWLEDGMENT}
% \vspace{-2mm}
% \noindent This material is based on work supported by the National Science Foundation grants NSF\#1846340, NSF\#2054744, and the GRFP DGE\#2139899. Any opinions, findings, and conclusions or recommendations expressed in this material are those of the author(s) and do not necessarily reflect the views of the National Science Foundation.

%%%%%%%%%%%%%%%%%%%%%%%%%%%%%%%%%%%%%%%%%%%%%%%%%%%%%%%%%%%%%%%%%%%%%%%%%%%%%%%%
% \vspace{-2mm}

\bibliographystyle{IEEEtran}
\bibliography{references}

\vspace{-1.5mm}
\section{APPENDIX}
\label{app:model_excerpt}

\noindent
\textit{Representative Quadcopter Model Excerpt.}
The following subset illustrates representative physical, performance, and
compatibility relationships from the finalized model. Additional constraints and constants
are omitted for brevity. Notation follows Sec.~\ref{notation_subsection}.

\begingroup
\small
\setlength{\abovedisplayskip}{2pt}
\setlength{\belowdisplayskip}{2pt}
\setlength{\jot}{1pt}

$B$, $M$, $E$, $FC$, $P$, $RX$, $PDB$, and $F$ denote the selected
battery, motor, ESC, flight controller, propeller, receiver,
power-distribution board, and frame, respectively. The objectives are
$\min J$, $\max v_{\max}$, and $\max m_H$.
\vspace{1mm}
\begin{align*}
m_0 &= B_m+4M_m+4E_m+FC_m+4P_m \\
    &\quad +RX_m+PDB_m+F_m,\\
J &= B_{\$}+4M_{\$}+4E_{\$}+FC_{\$}+4P_{\$} \\
  &\quad +RX_{\$}+PDB_{\$}+F_{\$},\\
A &= 4\pi(P_{\mathrm{diameter}}/2)^2,\\
P_h &= \frac{[(m_0+m_L)g]^{3/2}}
{\eta_h\sqrt{2\rho A}},\\
t_f &= \frac{\alpha B_C B_V}{P_h+P_a},\\
T_m &= \left[
\eta_h M_{P^{\mathrm{cont}}}
\sqrt{2\rho(A/4)}
\right]^{2/3},\\
m_H &= \frac{4T_m}{g}-m_0,\\
n_\ell &= \nu M_{k_V}B_V
\left(
1-\lambda\sqrt{\frac{(m_0+m_L)g}{4T_m}}
\right),\\
v_p &= \frac{\eta_p n_\ell P_{\mathrm{pitch}}}{60},\\
v_d &= \left[
\frac{2\eta_f(4M_{P^{\mathrm{cont}}}-P_h)}
{\rho C_D\kappa F_{\mathrm{wheelbase}}^2}
\right]^{1/3},\\
v_{\max} &= \frac{1}{2}
\left[
v_p+v_d-\sqrt{(v_p-v_d)^2}
\right].
\end{align*}

Here $v_p$ and $v_d$ are the pitch- and drag/power-limited forward
speeds, respectively, so the final expression selects their minimum.
The constants used above are
$m_L=0.300\,\mathrm{kg}$,
$\alpha=0.80$,
$\eta_h=0.55$,
$\eta_f=0.60$,
$\eta_p=0.35$,
$\nu=0.60$,
$\lambda=0.15$,
$C_D=1.10$,
$\kappa=0.20$,
$\rho=1.225\,\mathrm{kg\,m^{-3}}$,
$g=9.80665\,\mathrm{m\,s^{-2}}$, and
$P_a=5\,\mathrm{W}$.

\balance

\noindent
Representative constraints include
\begin{align*}
M_{V_{\min}} &\le B_V\le M_{V_{\max}},&
E_{I^{\mathrm{cont}}}B_V &\ge M_{P^{\mathrm{cont}}},\\
P_{\mathrm{diameter}}
&\le \frac{0.90F_{\mathrm{wheelbase}}}{\sqrt{2}},&
t_f &\ge 600\,\mathrm{s},
\end{align*}
\begin{align*}
4M_{P^{\mathrm{cont}}}
&\ge P_h+P_D,&
v_{\max} &\ge v_r=5\,\mathrm{m\,s^{-1}},
\end{align*}
where
\[
P_D=
\frac{\rho C_D\kappa F_{\mathrm{wheelbase}}^2v_r^3}
{2\eta_f}.
\]

\endgroup

\end{document}